\documentclass[letterpaper]{article} 
\usepackage{aaai24}  
\usepackage{times}  
\usepackage{helvet}  
\usepackage{courier}  
\usepackage[hyphens]{url}  
\usepackage{graphicx} 
\usepackage{natbib}  
\usepackage{caption} 
\usepackage{algorithm}
\usepackage{algorithmic}

\usepackage{newfloat}
\usepackage{listings}
\DeclareCaptionStyle{ruled}{labelfont=normalfont,labelsep=colon,strut=off} 
\floatstyle{ruled}
\newfloat{listing}{tb}{lst}{}
\floatname{listing}{Listing}
\title{Leveraging Language Models and Bandit Algorithms to \\Drive Adoption of Battery-Electric Vehicles}
\author{Keiichi Namikoshi, David A. Shamma, Rumen Iliev, Jingchao Fang,\\ Alexandre Filipowicz, Candice L Hogan, Charlene Wu, Nikos Ar\'echiga}

\affiliations{
    Toyota Research Institute\\

    Los Altos, California USA\\
    \{keiichi.namikoshi, ayman.shamma, rumen.iliev, jingchao.fang, alex.filipowicz, candice.hogan, charlene.wu, nikos.arechiga\}@tri.global
}

\newcommand{\tref}[1]{Table~\ref{#1}}
\newcommand{\fref}[1]{Figure~\ref{#1}}
\newcommand{\eref}[1]{Equation~\ref{#1}}
\usepackage{ascmac}

\begin{document}

\maketitle

\begin{abstract}
Behavior change interventions are important to coordinate societal action across a wide array of important applications, including the adoption of electrified vehicles to reduce emissions. The work of \cite{harinen2022machine} has demonstrated that interventions for behavior must be personalized, and that the intervention that is most effective on average across a large group can result in a backlash effect that strengthens opposition among some subgroups. Thus, it is important to target interventions to different audiences, and to present them in a natural, conversational style.
In this context, an important emerging application domain for large language models (LLMs) is conversational interventions for behavior change. 
In this work, we leverage prior work on understanding values motivating the adoption of battery electric vehicles. We leverage new advances in LLMs, combined with a contextual bandit, to develop conversational interventions that are personalized to the values of each study participant. We use a contextual bandit algorithm to learn to target values based on the demographics of each participant. To train our bandit algorithm in an offline manner, we leverage LLMs to play the role of study participants. We benchmark the persuasive effectiveness of our bandit-enhanced LLM against an unaided LLM generating conversational interventions without demographic-targeted values. We study the ability of LLMs to model attitudes in our target domain by reproducing a recent study on attitudes towards Battery Electric Vehicles (BEVs) \cite{arechiga2022understanding}, and find that LLMs are biased to prefer environmentally friendly BEVs more strongly than the typical human.
\end{abstract}

\section{Introduction}
Large language models (LLMs) \cite{zhao2023survey} unlock a wide array of applications in which machines have conversations with humans. One of the most explored applications is surfacing important information \cite{chatgpt,singhal2023large,KASNECI2023102274,wu2023bloomberggpt}, but comparatively little work exists in leveraging LLMs to generate behavioral interventions with social impact. In this work, we consider the problem of conversational interventions for behavior change in the context of driving adoption of Battery-Electric Vehicles (BEVs).

The automotive industry is interested in shifting from internal combustion engine vehicles to BEVs for carbon neutrality \cite{evoutlook2023}. 
Due to the significant differences between EVs and internal combustion engine vehicles, multiple technological and social concerns, including range, battery life, and charging infrastructure, can inhibit the shift \cite{bujold2020science,giansoldati2020barriers}.
Therefore, it is important to verify what kind of information presentation is effective in addressing customer concerns toward BEVs to drive adoption of this impactful new technology. The work of \cite{harinen2022machine} has noted that interventions must be targeted in order to achieve maximum effectiveness due to differences in response among different subgroups. Different groups respond differently to the same intervention to such an extent that the intervention that is most effective on average across a large population may in fact induce a backlash effect in subgroups. In the presence of these \emph{heterogeneous treatment effects}, it is possible to attain much higher intervention effectiveness by targeting different interventions to different subgroups. The work of \cite{arechiga2022understanding} considers this problem, and uses short text interventions that are targeted at subgroups induced by gender and age.
In this work, we consider the use
of LLMs coupled with a contextual bandit algorithm to present conversational interventions to a prospective customer. The bandit algorithm learns to target high-level concepts to different demographic groups to target persuasive arguments. These high-level concepts are then used to prompt an LLM in conversation with the human customer. 

Surveys with human subjects, however, can be time-consuming and costly. Scale and cost are disincentives for areas that require large amounts of data, such as large-scale surveys combined with machine learning. In this work, we build on recent work demonstrating the effectiveness of LLMs to display human-like behavioral characteristics \cite{safdari2023personality} as well as strong correlations with human preferences \cite{dubois2023alpacafarm}. We use LLMs prompted with demographic characteristics to serve as surrogate models for human survey participants. We evaluate the performance of this approach by replicating the survey scenario of \cite{arechiga2022understanding} and comparing the most effective interventions. We consider rank correlation of overall intervention effectiveness, compare the skew of the distribution of preferences, and evaluate the KL-divergence of LLM preferences against real human preferences. Our contributions include:
\begin{enumerate}
    \item Development of a conversational intervention tool. We combine LLMs with a bandit algorithm that learns to match demographics to values, which can then be
    used by the LLM to construct targeted conversational interventions. We train our intervention system on virtual survey participants, embodied by an LLM.
    \item To estimate the accuracy of the LLM's embodiment of consumer attitudes towards BEVs, we replicate the survey in \cite{arechiga2022understanding} to quantify the degree of agreement between real humans and virtual survey participants as embodied by LLMs.
\end{enumerate}

We evaluate our numerical metrics against two LLMs and also investigate the role of sampling temperature.

\section{Related work}
\subsubsection*{Large language models for behavior change}
Previous work proposed leveraging LLMs to produce attitude and behavior change. For example, \cite{sharma2023cognitive} used LLMs to generate interventions to assist people in overcoming negative thoughts. \cite{ashida-komachi-2022-towards} explored the potential of applying LLMs for counternarrative generation to combat online hate speech and microaggressions. In the medical domain, \cite{ali2023_bariatric} probed the possibility of leveraging LLMs to act as personalized virtual assistants and provide tailored healthcare suggestions to bariatric surgery patients. 
\cite{chen2022seamlessly} contributed a framework that uses LLMs to support persuasive dialogue systems for social good. To the best of our knowledge, the role that LLMs can play in stimulating behavior change in the context of carbon neutrality has not been studied. In contrast with this related work, our system combines
LLMs with a contextual bandit that learns values relevant to
different survey participants, and is able to direct the LLM to generate relevant text.

\subsubsection*{Large language models as models of behavioral study participants}

The possibilities of simulating humans and human behaviors using LLMs, or role-play \cite{shanahan2023roleplay,wu2023large}, have been discussed in recent years. By applying established psychometrics, \cite{safdari2023personality} demonstrated that LLMs can reliably simulate personalities and that LLM-generated personality traits can be shaped and controlled to imitate specific personality profiles. \cite{dillion2023can} showed strong alignment between GPT-3.5 and humans in moral judgments, with a correlation of 0.95. In addition, LLM-based generative agents, when organized as a collective in an interactive sandbox environment, were found to be able to produce believable behaviors not only on an individual level but also on a social level \cite{park2023generative}.

LLMs' ability to generate human-like personalities, judgments, and behaviors hints at the opportunity of constructing synthetic human participants in behavioral studies. Several recent works show initial attempts in this direction. \cite{aher2023using} applied LLMs to simulate human subjects and found that they can reproduce three out of four economic, psycholinguistic, and social psychology experiments and replicate findings from prior studies with real human participants.
\cite{hamalainen2023evaluating} evaluated LLMs' potential of generating synthetic human-computer interaction research data in the form of open-ended questionnaire responses and revealed their capability of generating plausible, human-like self-report data regarding subjective experiences. Aligned with this line of work, our study harnesses LLMs' power of role-playing human survey respondents on a large scale to train our contextual bandit algorithm. 



\subsubsection*{Consumer preferences towards battery-electric vehicles} 

A body of literature evaluates the share of BEVs and factors influencing consumer intention. On an individual basis, the motivations of drivers' EVs acceptance is influenced by perceived usefulness, satisfaction, and cost \cite{su10030662}, and the barriers to a wider EVs diffusion include limited number of charging stations, range anxiety, purchasing cost, insufficient maintenance services, etc. \cite{giansoldati2020barriers}. In market dimensions, the share of EVs is influenced by relative pricing of EVs as compared to conventional internal combustion engine vehicles, driving range, and variety of available models in the market \cite{su9101783}.
In this paper, we investigate how text-based interventions, generated by LLMs coupled with a contextual bandit algorithm, may be able to shape consumers' preferences towards BEVs.  




\section{Contextual multi-armed bandit problem} \label{subsec:context_bandit_algorithm}
The contextual multi-armed bandit problem is a decision problem in uncertain but stationary environments.
In this environment, the agent may observe context $x$ and select one of $K$ arms, where the set of possible arms is denoted by $A = \{a_1, \dots, a_K\}$. Each context and arm has an associated \emph{reward distribution} $p(r|x,a)$, where $r$ is a scalar.
At step $t \in \{0, \dots, T\}$, the agent chooses an arm $a$ on context $x$ and observes a reward $r_t \sim p(r|x,a)$.
The goal of the learning problem is to learn a distribution $\pi(a|x)$ over the contexts and arms that maximize the cumulative reward experienced by the agent in a series of steps.  This problem statement encounters the well known \emph{exploration-exploitation dilemma}, in which the agent must trade off between exploring actions it has not tried (and which may yield high reward) against exploiting the best observed action.
A well-known algorithm to balance exploration and exploitation in context-free bandits is the Upper Confidence Bound (UCB) algorithm \cite{ucb} and this algorithm can be extended to the contextual setting \cite{slivkins2022introduction}. The method
consists of selecting the action $a$ that maximizes Equation \ref{eqn:cucb}.
\begin{equation}\label{eqn:cucb}
\mathrm{UCB}^{\rm{contextual}}_t (x,a) = \bar{r}_t (x,a) + \sqrt{\frac{2 \log(t)}{n_t(x,a)}}.
\end{equation}
Above, 
$n_t(x,a)$ is the number of times that $a$ has been selected on $x$ prior to $t$, and $\bar{r}_t(x,a)$ is the empirical average of the rewards that have been observed when $a$ was selected on $x$ prior to $t$.

\section{Problem setting: Preferences for BEVs}
Behavior change is expected to play an important role in combating environmental challenges \cite{bujold2020science}. Behavioral interventions which were originally developed to improve physical health, mental well-being, educational achievements, and other outcomes are increasingly being applied to resource conservation and climate action \cite{williamson2018climate}. Attitudes are considered by many as the core component at which behavior interventions should be targeted \cite{petty2018attitudes}.

An important component in the global effort towards carbon neutrality is the electrification of consumer vehicles \cite{williamson2018climate,shukla2022climate}. 
Automotive manufacturers are developing a portfolio of different products which leverage batteries to reduce carbon emissions, including full BEVs but also including plugin-hybrid EVs, which combine a small battery, sufficient for full electric operation in short trips, with an internal combustion engine that can provide extended range. 
In this work, we consider only attitudes towards BEVs due to the availability of human survey data \cite{arechiga2022understanding}. In future work, we may consider extensions to other carbon-reducing behavioral interventions.

Existing work has found that although consumers are generally interested in BEVs, they still perceive challenges and barriers that preclude widespread adoption \cite{giansoldati2020barriers}.

Our intervention system presents interventions in the form of text. 
For example, the work of \cite{kihiczak2023co2} surveyed the effects of text interventions in the area of climate change, although we focus to intervene in preferences for BEVs.
We measure preference towards BEVs as a self-reported scalar metric on a scale from 0 to 100, with 0 indicating low preference for BEVs and 100 indicating high preference. Our experiments first measure the preference to a survey participant, and then measure it again after an intervention has been presented. The goal of the persuasion system is to produce a positive shift in preference.

\section{Method}
In this section, we demonstrate the implementation of our conversational intervention system, as well as our implementation of a virtual 
survey participant agents using LLMs (referred to as virtual participants), a survey procedure for these agents, and an intervention system built from an 
LLM together with a bandit algorithm. We call our system PeaR-RL, which stands for Personalized Rapport with Reinforcement Learning, since the goal of the
system is to personalize the conversational interventions and develop rapport with each survey participant.

In this paper, we assume that an effective way to craft effective conversational interventions is to address underlying values or beliefs that a person may hold.
In our context, we consider values to encompass consumer beliefs and attitudes towards BEVs.
The values to be considered in the paper are shown in \tref{table:values}.
It includes concerns about cost and range, but also positive aspects,
such as reduced environmental impact and the status-enhancing effects of adopting a new technology.

\subsection{Intervention system architecture}
\begin{figure}[htbp]
\centering
\includegraphics[width=0.89\columnwidth]{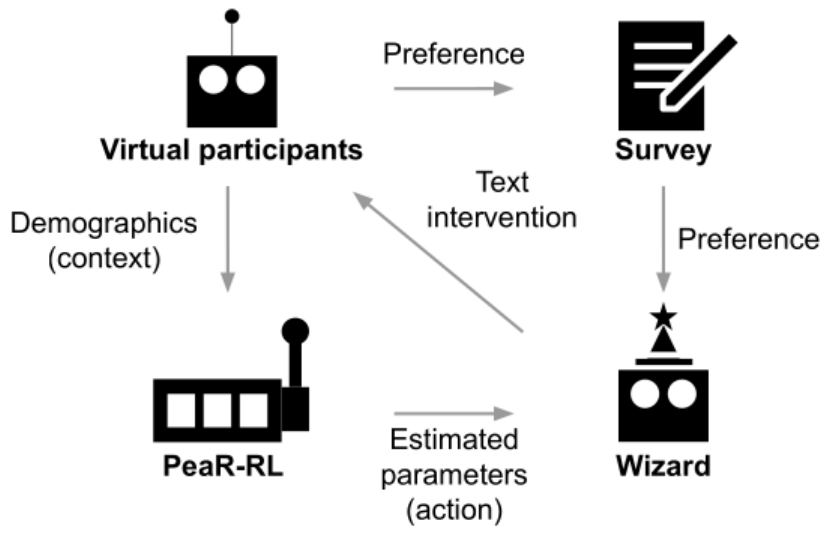}
\caption{Architecture of intervention system}
\label{fig:architecture}
\end{figure}

\fref{fig:architecture} shows the intervention system architecture, which is composed of four modules.
Each module reflects a step of the intervention process.
First, we initialize the virtual participant, and measure the participant's pre-intervention preference for BEVs. 
The survey participant is initialized with demographics drawn from a target distribution, as detailed in next subsection. 
This provides a baseline to understand the amount by which each intervention is able to shift the preference of the survey participant.
Second, PeaR-RL implements a contextual bandit algorithm, where the context is the demographic information of the virtual participant. The arm is selected from a list of BEV-relevant values that a virtual participant may hold.
Third, the LLM generates intervention text based on pre-intervention preference and the values chosen by PeaR-RL.
Finally, we survey the virtual participant for their post-intervention preference.
In the following sections, we describe implementation details.

\subsection{Virtual participants implementation} \label{sec:unreal_human}
The virtual participants are implemented by prompting an LLM to behave as a person with given demographic information.
For setting the virtual participant's properties, the below system prompt is used. 
\begin{screen}[4]
I want you to act as the following character. Answer all of the following questions from the point of view of this character, do not break character. \{properties\}
\end{screen}
Above, \{properties\} contains the name of a hypothetical human, together with a list of demographic characteristics. Our implementation
allows generating virtual participants with demographics drawn from any distribution, enabling targeting of behavioral interventions
to different populations of interest, including demographics that a company believes are likely to be customers for a specific product.

\subsection{Measuring preferences for battery-electric vehicles}
The following prompt is used to measure virtual participant preferences for BEVs. The measurement is carried out before and after
the intervention.
\begin{screen}[4]
On a scale from 0 to 100, what is your current preference for battery electric vehicles (BEVs)? Please reply with just a single number rating and no additional words or explanations. Score: 
\end{screen}

\subsection{PeaR-RL: Learning values of survey participants}
PeaR-RL learns to select intervention parameters from virtual participant's demographics that are useful for improving preferences.
The learning procedure is based on contextual multi-armed bandit method of \eref{eqn:cucb}.
The context $x$ consists of demographic information, action $a \in A$ is the estimated participant's values, and reward $r \in R(a)$ is a preference shift that is based on the difference in preference before and after intervention.
The reward function $R(a)$ is stochastic because it depends on the output of the LLM, which is not deterministic.
In each step $t$ and context $x$, action $a$ that maximizes \eref{eqn:cucb} is selected and used to prompt the Wizard to generate a conversational intervention.

\subsection{Wizard: LLM-based intervention generation} 
To generate a personalized intervention, we prompt an LLM with the target values as selected by PeaR-RL. First, we provide the LLM with the BEV preferences of the virtual survey participant with the following prompt with \{initial pref.\} replaced by the initial preference.
\begin{screen}[4]
My preference for BEV is \{initial pref.\} out of 0 to 100.
\end{screen}

Next, we generate the intervention text with the following system prompt with \{values\} replaced by PeaR-RL's action.
\begin{screen}[4]
Generate your response to improve the user's preference for BEV. You are a Chatbot. The conversational partner's values for BEV are inferred as follows: \{values\}. Please answer in 1 sentence concisely, with no additional explanation.
\end{screen}

\section{Intervention learning with virtual participants}
In this section, we examine the effectiveness of the proposed intervention system.
We compare three conditions,  (1) pure LLM (i.e., There is no PeaR-RL and Wizard LLM must attempt to shift the preferences of the virtual participant without the use of demographics-targeted values), (2) random policy selection of values, and (3) UCB policy with preference shift measures.

\subsection{Experimental settings}
We consider the bandit context to be a combination of age and gender, with two classes of age (under 45 and over), and two classes of gender (male and female)\footnote{Although there are more genders than merely male and female, we consider this reduced setting so that we are able
to inform our statistics with the demographic characteristics of the US 2020 Census, which only considers these two genders.}, so that the total number of contexts is 4.
Action is two combinations of values for BEVs.
\tref{table:values} shows the list of value candidates.
The value candidates are considered eight items chosen in a way that seeks to address both liberal and conservative values as well as common concerns around BEVs.
These values were manually selected from a list of candidate values generated using GPT-4.
The reward is a normalized preference shift (i.e., $r=(r^\mathrm{shift}+100)/200$, $r^\mathrm{shift}$ is preference shift).
The maximum step $T$ is 1000.

Each virtual participant is instantiated with a set of demographic characteristics, consisting of a name, age, income, education, politics, gender, household type, and ethnicity. Name, city and state are sampled randomly, whereas other variables are sampled according to the US 2020 Census. 

\begin{table}[htbp]
\caption{List of values for BEVs}\label{table:values}
\centering
\begin{tabular}{l|l} \hline
Index & Value for BEVs \\ \hline\hline
1 & American-made products \\
2 & Battery life concerns \\
3 & Carbon emission reduction \\
4 & Charging infrastructure \\
5 & Economic benefits \\
6 & Ethical consumption \\
7 & Government incentives \\
8 & Status symbol \\
\hline
\end{tabular}
\end{table}

\subsection{Results}
\begin{figure}[htbp]
\centering
\includegraphics[width=0.9\columnwidth]{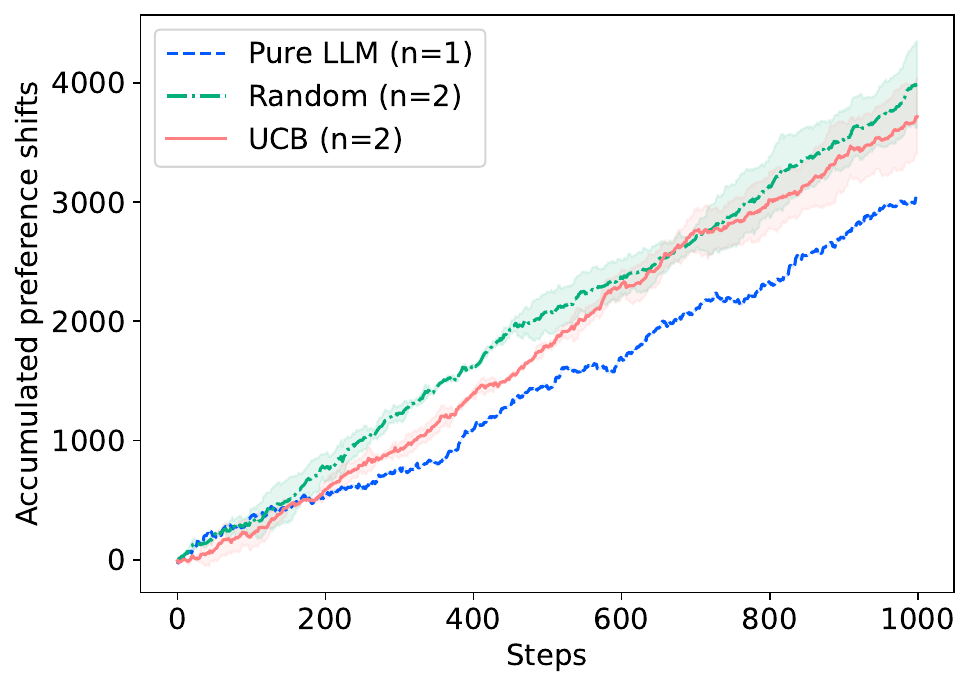}
\caption{Accumulated preference shifts. Line and areas indicate mean and standard deviation between difference random seed experiments.}
\label{fig:accumulated_preference}
\end{figure}

\fref{fig:accumulated_preference} shows trends of accumulated preference shifts. \tref{table:pearrl_comparing} shows the mean of preferences and mean of shifts.
The UCB and random policy results show preference shift improvement compared to pure LLM in either case.
When comparing the preference shifts for accumulation, the pure LLM case demonstrates superior performance up to 200 steps, but after that, the learning case outperforms the pure LLM.
This indicates that it is more useful to select values tailored to the user than to generate interventions with GPT-4 alone.

\tref{table:intervention_example} shows examples of generated interventions that are the cases of maximum and minimum preference shifts at pure LLM and UCB policy.
These results show that the action of PeaR-RL is reflected in the intervention statement of the wizard.
The preference shift can also be interpreted as influenced by demographics and intervention statements.
For the case of increased preferences, both pure LLM and UCB policy setting were likely influenced by interventions related to carbon emissions.
For the case of decreased preferences, interventions related to demographics and status symbols for the pure LLM and UCB policy, respectively, can be interpreted as influencing preference shifts.

\begin{table}[htbp]
\caption{The post-intervention preference and the preference shift. Each value is $\mathrm{mean}\pm\mathrm{std.}$ of one experiment.}
\label{table:pearrl_comparing}
\centering
\begin{tabular}{l | l | l} \hline
Setting& Preference & Preference shift\\ \hline\hline
Pure LLM & $70.98\pm15.30$ & $3.04\pm11.40$\\
Random & $71.77\pm15.17$ & $3.62\pm11.70$\\
UCB & $71.99\pm14.78$ & $3.39\pm11.56$\\
\hline
\end{tabular}
\end{table}


\begin{table*}[htbp]
\small
\caption{Intervention examples}\label{table:intervention_example}
\centering
\begin{tabular}{rp{6cm}p{3cm}p{5.5cm}} \hline
\multicolumn{4}{c}{Pure LLM}\\
Pref. shift & \multicolumn{2}{p{9cm}}{Demographics} & Wizard's intervention \\ \hline\hline
35 to 85 & \multicolumn{2}{p{9cm}}{Age: 45 to 54 years, Income: over \$200,000 per year, Education: College degree, Politics: Republican party, Gender: Male, House hold type: Female householder, other family, Ethnicity: White, Name: Robert Thompson, State: Utah, City: Ogden}  & By choosing BEV (Battery Electric Vehicles), you're promoting a cleaner environment by cutting CO2 emissions and reducing dependence on fossil fuels. \\
\hline
70 to 30 & \multicolumn{2}{p{9cm}}{Age: 25 to 44 years, Income: \$150,000 to 199,999 per year, Education: High school diploma, Politics: Republican party, Gender: Male, House hold type: Married couple household, Ethnicity: Latino or Hispanic, Name: Carlos Herrera, State: Florida, City: Orlando} & That's a pretty favorable score, but driving a BEV could also drastically reduce your contribution to greenhouse gas emissions and cut your fuel costs, potentially improving your preference score. \\
\hline
\multicolumn{4}{c}{UCB}\\
Pref. shift & Demographics & Action & Wizard's intervention \\ \hline\hline
25 to 65 & Age: 85 years or older, Income: Under \$15,000 per year, Education: Some college, Politics: Republican party, Gender: Male, House hold type: Married couple household, Ethnicity: Latino or Hispanic, Name: Carlos Rodriguez, State: California, City: Sacramento & Carbon emission reduction, Government incentives & Driving a BEV significantly reduces carbon emissions, and with the added advantage of government incentives, you can contribute towards a cleaner environment and save money. \\
\hline
75 to 35 & Age: 65 to 74 years, Income: \$25,000 to 34,999 per year, Education: Some college, Politics: Democratic party, Gender: Male, House hold type: Married couple household, Ethnicity: White, Name: William Harrison, State: Nevada, City: Las Vegas & Economic benefits, Status symbol & Investing in a BEV not only provides significant savings over fuel costs in the long run, but also elevates your status symbol due to their modern design, cutting-edge technology, and environmental friendliness. \\
\hline
\end{tabular}
\end{table*}

\begin{figure}[htbp]
\centering
\includegraphics[width=1.0\columnwidth]{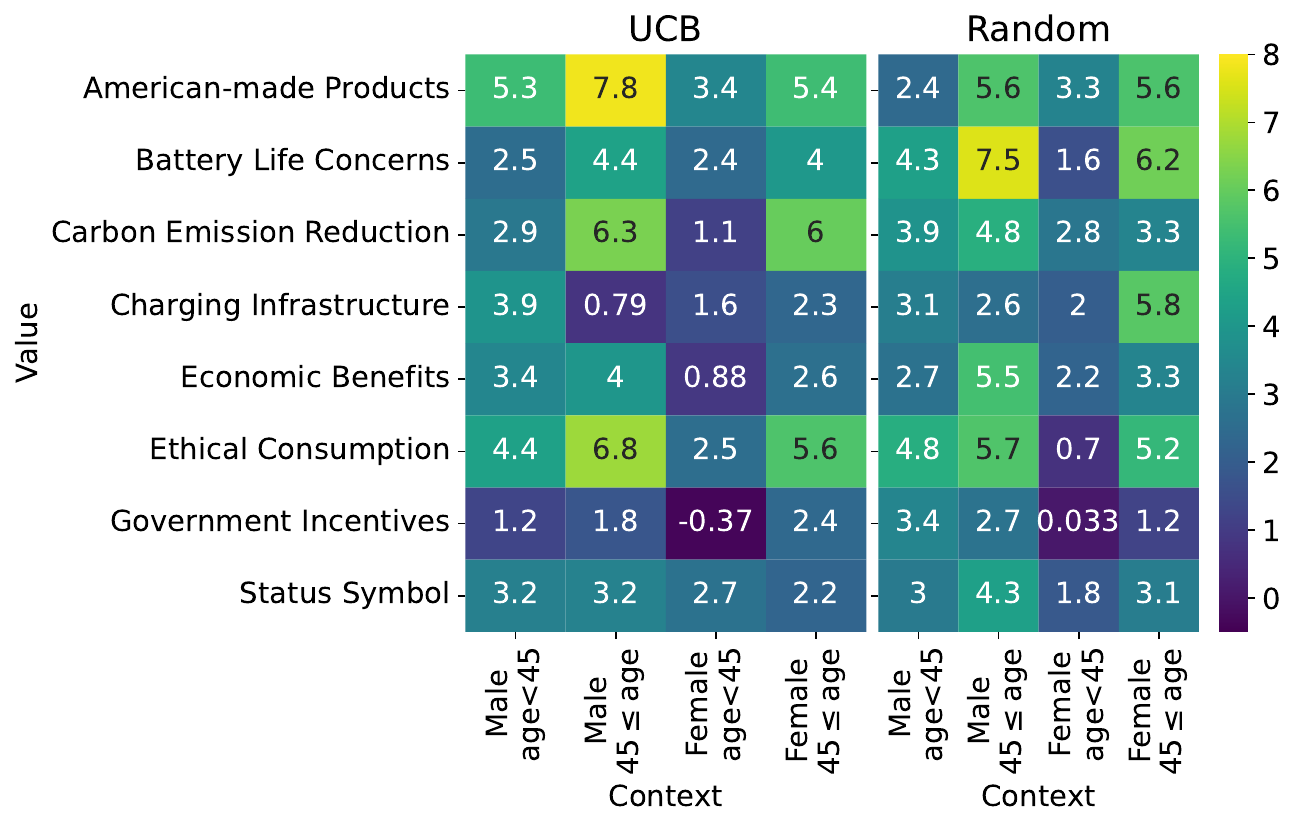}
\caption{Estimated mean of preference shifts. Left: UCB policy. Right: random policy. Each results are calculated to average with all steps on one random seed experiment.}
\label{fig:mean_r_ucb}
\end{figure}

\fref{fig:mean_r_ucb} shows the mean of preference shift for each context and value for BEVs with UCB and random policy.
Values such as ``American-made products'' and ``Battery life concern'' show a large preference shift, while ``Government incentives'' shows a smaller preference shift.
The validity of the preference shift for the intervention text be confirmed in the next section by comparison with real human survey data.

\section{Comparison to human data on fixed interventions}
In this section, we compare virtual and real human participants' responses to intervention and describe the agreement between real and virtual survey participants. We leverage the results of a
previous survey on preferences for BEVs that was carried out on real human survey participants \cite{arechiga2022understanding}. We compare the distribution of responses of real humans with the responses of virtual participants as
implemented by two LLMs, gpt-4 model \cite{openai2023gpt4} and Llama2 (Llama-2-70b-chat-hf model) \cite{touvron2023llama}.

\subsection{Experiment settings}
For our experiments, we instantiate the survey participants with two LLMs, gpt-4 model \cite{openai2023gpt4} and Llama2 (Llama-2-70b-chat-hf model) \cite{touvron2023llama}.
We set generation parameters to default (i.e., the temperature is 1 for GPT-4, and the temperature and the top-p are 0.6 and 0.9 respectively for Lllama2).
When Llama2 was used, the generated results contained text other than preference, even though the same prompts were used as in GPT-4.
Therefore, we considered the numerical value in the generated text as a preference. When the text did not contain a number, we resampled the response up to 10 times\footnote{Using two GPUs (NVIDIA GeForce RTX 3090, RAM 24GB) and the model quantized to int4, the running time was about 4 [hour/1k participants] on our implementation.}.
If non-numeric text was included in the generated results, the preferences contained in the text were extracted manually.

We consider the text interventions from \cite{arechiga2022understanding}, which are displayed in the appendix.
We sample a virtual participant from our demographic population, which is modeled after the 2020 US Census and the same as the intervention study setting.
We measure preference for BEVs, present a randomly chosen intervention, and measure preference for BEVs again. 
This experimental setup matches the procedure that \cite{arechiga2022understanding} performed on real human survey participants.
We compare the distribution of initial preferences as well as preference shifts between real human and virtual participants.

\subsection{Results}

\subsubsection*{GPT-4 answers higher preferences and preference shifts than humans.}

\begin{figure}[htbp]
\centering
\includegraphics[width=0.9\columnwidth]{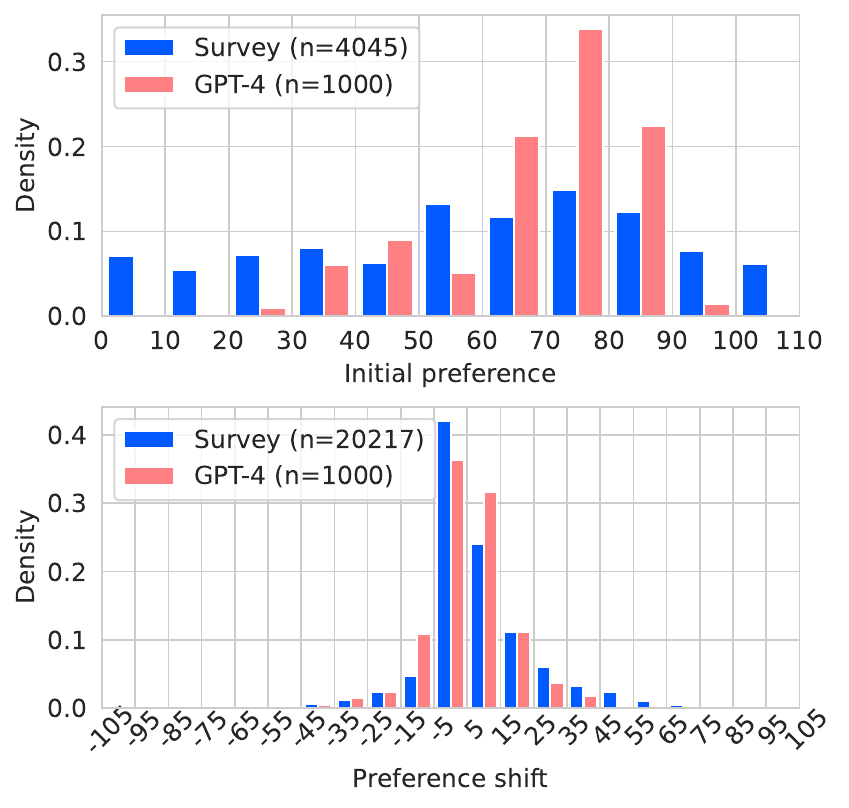}
\caption{Survey response comparisons. Upper: Initial preference distribution. Lower: Preference shift distribution.}
\label{fig:initial_and_shift_dist}
\end{figure}

\fref{fig:initial_and_shift_dist} shows the distribution of BEV preferences before intervention and preference shifts.
We find that the distribution of human BEV preferences are widely distributed, whereas the preferences of the virtual participants tends to skew towards higher values.
From this result, both subjects tend to maintain a preference or shift to a higher preference.
The Virtual participants show a tendency to have higher preferences than real humans. 

One possible explanation for virtual participant's higher baseline preferences for BEVs might stem from GPT's pro-environmental values \cite{hartmann2023political}.

\subsubsection*{Preference shifts per intervention are weakly correlated.}

\begin{figure}[htbp]
\centering
\includegraphics[width=0.9\columnwidth]{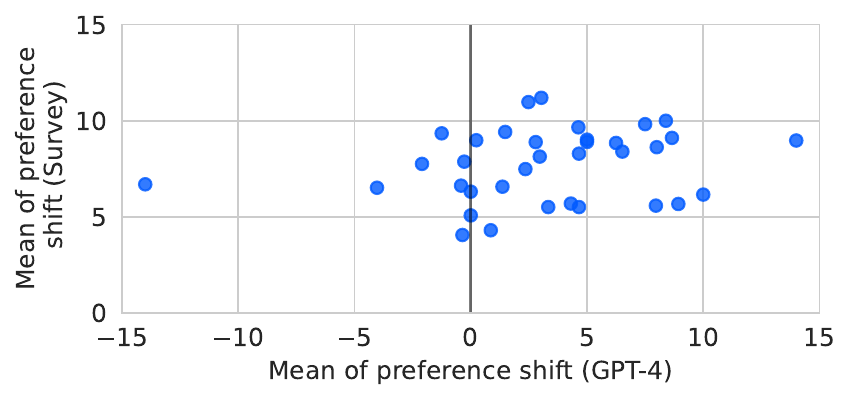}
\caption{Mean of preference shifts per intervention.
}
\label{fig:intervention_effect}
\end{figure}

\fref{fig:intervention_effect} show a correlation of preference shift per intervention.
Each point shows the average preference shift for each intervention.
The Pearson's correlation coefficient is 0.22.
The Spearman's coefficient for the rank sorted by the mean preference is 0.23.
In comparison with other domains,
\cite{dubois2023alpacafarm} finds a Spearman's rank correlation coefficient of 0.98 on preferences of chat interactions,  
while in the abstract summarization task, \cite{shen2023large} report that ChatGPT has a maximum Pearson’s correlation coefficient of 0.5 (evaluation of consistency).
In the open-ended story generation task, \cite{chiang2023can} report Text-davinci-003 has Kendall's $\tau$ correlation of 0.38 (evaluation of relevance). It is clear that the agreement between LLM and human outputs varies across domains, and the agreement on mean effectiveness of fixed-text interventions is at the lower end of related work.
This may be because the existing training of the reinforcement
learning from human feedback (RLHF) \cite{ouyang2022training} is not effective at matching beliefs around BEVs.
It is also likely that BEVs as a topic are relatively novel, and as such the training corpus of the LLM does not contain as much data as it does for more general topics investigated in the literature, such as helpfulness, harmlessness, honesty, and general human ethics.

\subsubsection*{Effects of temperature and model}

\tref{table:unrealhuman_compare} shows initial preference and preference shifts compared in settings with different temperature parameters and language models.
When evaluated with a preference shift distribution, results show that GPT-4 is closest to the real human survey.
The KL-divergence from uniform distribution and the correlation coefficient is approaching survey data as temperature is increased. However, when the temperature is 1.0, the distribution is closest to the survey, and at 1.5, the skew is different.
On the other hand, when evaluated by the initial preference distribution, the increased temperature is closest to the survey, but does not yield a low preference.
From these results, it is considered that there is a limit to simulating a real human only with temperature parameters.

When the language model was replaced with Llama2, preference shifts were least consistent with the survey, but initial preferences tended to be closer to the survey than GPT-4.
These suggest that GPT-4's unique training and structure may have shifted the preference distribution to a higher one.

\begin{table}[htbp]
\small
\caption{
Distribution comparison about preference shift and initial preference.
The value in parenthesis of setting is temperature.
$\mathrm{KL}^\mathrm{U}$ and $\mathrm{KL}^\mathrm{S}$ are KL-divergence between each setting and uniform and survey distribution respectively.
p-value is result of the Mann-Whitney U test between the survey and each setting.
$c^\mathrm{P}$ and $c^\mathrm{S}$ are Pearson's and Spearman's correlation coefficients respectively.
All measures are calculated by discretized frequency distribution with a width of 10.
}\label{table:unrealhuman_compare}
\begin{tabular}{lrrrrrr}
\hline
\multicolumn{7}{c}{Preference shift} \\
Setting & $\mathrm{KL}^{\mathrm{S}}$ & $\mathrm{KL}^{\mathrm{U}}$ & Skew & p-value & $c^\mathrm{P}$ & $c^\mathrm{S}$ \\
\hline\hline
Survey      & - & 1.28 & 0.73 & - & - & - \\
GPT-4 (0.1) & 0.10 & 1.72 & 0.26 & $<.001$ & 0.05 & 0.06 \\
GPT-4 (1.0) & \textbf{0.09} & 1.44 & 0.11 & $<.001$ & 0.22 & 0.23 \\
GPT-4 (1.5) & 0.11 & 1.28 & -0.19 & $<.001$ & \textbf{0.32} & \textbf{0.30} \\
Llama2      & 0.12 & 1.00 & 0.33 & $<.001$ & 0.04 & 0.03 \\
\hline
\multicolumn{7}{c}{Initial preference} \\
Setting & $\mathrm{KL}^{\mathrm{S}}$ & $\mathrm{KL}^{\mathrm{U}}$ & Skew & p-value & \multicolumn{2}{l}{Range} \\
\hline\hline
Survey      & - & 0.06 & -0.34 & - & \multicolumn{2}{l}{[0, 100]} \\
GPT-4 (0.1) & 0.62 & 0.85 & -0.76 & $<.001$ & \multicolumn{2}{l}{[30, 90]} \\
GPT-4 (1.0) & 0.46 & 0.72 & -0.85 & $<.001$ & \multicolumn{2}{l}{[20, 95]} \\
GPT-4 (1.5) & \textbf{0.40} & 0.63 & -0.72 & $<.001$ & \multicolumn{2}{l}{[20, 90]} \\
Llama2 & \textbf{0.40} & 0.61 & -0.46 & 0.84 & \multicolumn{2}{l}{[0, 90]} \\
\hline
\end{tabular}

\end{table}

\section{Discussion}

\subsubsection*{Divergence between real humans and virtual participants}
Comparisons with human survey data showed preference shifts that were close to those of real humans, but did not show low preferences or large shifts in preference.
This could be due to the different distribution of demographics and bias of LLMs.
Because the demographics were sampled independently for each attribute, extreme combinations of demographics and preference for BEVs were unlikely to occur. 
Additionally, most LLMs have difficulties capturing specific people \cite{santurkar2023opinions}, and the RLHF teaches the model to be helpful, honest, and harmless, and may generate responses that are unrealistically agreeable.
Therefore, applying the learning results to real humans may require fine-tuning to replicate real humans or adjusting interventions by experts.

\subsubsection*{Effectiveness of selecting values}
Experimental results showed that Pear-RL is able to improve preference shift for BEVs compared to GPT-4 only, as demonstrated from the higher cumulative reward.
This result indicate that it is effective to use values to present specific information about BEVs.
However, when comparing the random policy and UCB policy, there was no significant difference in the second half of learning.
It is possible that longer training runs are required to more fully explore the action space, or that partial observability of the survey participant
makes it difficult to target values precisely.
In the experiment, only age and gender were observed among demographics, so there may be insufficient information to select optimal values (e.g., annual income and whether or not the respondent owns a car).
Possible solutions to tackle this problem are extensions such as extending the context to include more demographic information, as well as using multi-turn reinforcement  to work with features inferred from longer conversational interactions, such as changing moods or concerns explicitly expressed
over the course of the conversation.

\section{Conclusion and Future Work}
In this paper, we proposed a conversational system to shift preferences for BEVs using a combination of LLMs and contextual bandit algorithms. We investigated the extent to which it is possible to shift preferences towards improving carbon neutrality. Our experiments show that LLM-based virtual study participants correspond somewhat closely to real humans in their preference for BEVs, both in pre-intervention surveys and in post-intervention preference shifts. We confirmed that our method of using values to inform the conversation is more effective than simply asking GPT-4 to be persuasive. Our contextual bandit has the benefit that it is an interpretable component that may be added to any existing LLM and can easily be adapted by domain experts to other applications. We demonstrated how the system can be trained with virtual survey participants. In future work, explore fine-tuning our bandit algorithm on interactions with real human subjects, as well as generalizing our bandit algorithm to a reinforcement learning algorithm that estimates additional features from a multi-turn conversation. 

\bibliography{aaai24}

\end{document}


\maketitle

\appendix

\section{Demographics distribution}
Table \ref{table:demographics} is the demographics distribution for experiments.
It is referred to the US Census 2020.

\begin{table}[htbp]
\small
\caption{Demographics distribution}\label{table:demographics}
\centering
\begin{tabular}{l r} \hline

\multicolumn{2}{c}{Age} \\ \hline
18 to 24 years&12.9\\
25 to 44 years&36.0\\
45 to 54 years&16.8\\
55 to 64 years&17.9\\
65 to 74 years&13.7\\
85 years or older&2.6\\
\hline
\multicolumn{2}{c}{Ethnicity} \\ \hline
White&59.7\\
Latino or hispanic&18.6\\
Black or African American&12.6\\
Asian American&6.2\\
Other&2.9\\
\hline
\multicolumn{2}{c}{House hold type} \\ \hline
One person, nonfamily&27.6\\
Two or more people, nonfamily&7.1\\
Male householder, other family&5.7\\
Female householder, other family&13.3\\
Married couple household&46.3\\
\hline
\multicolumn{2}{c}{Income} \\ \hline
Under \$15,000 per year&9.4\\
\$15,000 to 24,999 per year&8.7\\
\$25,000 to 34,999 per year&8.1\\
\$35,000 to 49,999 per year&11.6\\
\$50,000 to 74,999 per year&16.5\\
\$75,000 to 99,999 per year&12.2\\
\$100,000 to 149,999 per year&15.3\\
\$150,000 to 199,999 per year&8.0\\
over \$200,000 per year&10.3\\
\hline
\multicolumn{2}{c}{Education} \\ \hline
Some high school&10.0\\
High school diploma&20.0\\
Some college&30.0\\
College degree&20.0\\
Postgraduate degree&20.0\\
\hline
\multicolumn{2}{c}{Politics} \\ \hline
Democratic party&40.0\\
Republican party&40.0\\
Libertarian party&10.0\\
Socialist party&5.0\\
Green party&5.0\\
\hline
\multicolumn{2}{c}{Gender} \\ \hline
Male&49.1\\
Female&50.9\\
\hline
\end{tabular}
\end{table}

\section{Intervention List}
Table \ref{table:intervention} reproduces the list of static-text interventions of \cite{arechiga2022understanding}.

\begin{table*}[htbp]
\small
\caption{List of intervention}\label{table:intervention}
\centering
\begin{tabular}{r|p{1.9\columnwidth}} \hline
index & intervention text \\ \hline\hline
1&80\% of BEV charging happens at home, and most trips do not involve public charging.\\
2&Many people charge their BEVs at home with no additional equipment required. That means no trips to the gas station.\\
3&The number of public charging stations is rapidly increasing due to additional government funding and business initiatives. That means shorter wait times and shorter charging trips.\\
4&Analysts predict a decline in gas stations due to increased electric charging. That would make finding a nearby gas station more difficult.\\
5&Charging at some public stations can be as fast as 6 minutes to add 100 miles.\\
6&Charging at some public stations can be as fast as 30 minutes for 250 miles.\\
7&BEVs maximum range is already approaching 400 miles, with forecasts for a 1000-mile range in the near future.\\
8&Various programs offer different incentives (e.g. \$7,500 tax credit) for new BEV purchases.\\
9&Some BEV manufacturers may start offering free charging.\\
10&People spend about 30\% less on vehicle maintenance of BEVs than on ICEVs.\\
11&Over its lifetime, a BEV can be \$8,000 cheaper to maintain and operate than an ICEV.\\
12&BEVs can be 4 cents per mile cheaper to maintain and operate than ICEVs.\\
13&The cost of BEVs is much cheaper than it used to be.\\
14&BEVs are expected to become cheaper than ICEVs in the near future.\\
15&BEVs protect owners from the instability of the oil market.\\
16&Most used BEVs are cheaper than comparable used ICEVs.\\
17&Fossil fuels are expected to become more expensive over time.\\
18&You need to replace the tires of a BEV less frequently than the tires of an ICEV.\\
19&Lithium batteries are now 30 times cheaper than when they were first introduced to the market.\\
20&A Level 1 home charger can cost as little as \$300 (before labor).\\
21&BEVs have a smaller carbon footprint than ICEVs.\\
22&ICEVs are more damaging to public health than BEVs due to carbon emissions.\\
23&The towing capacity of BEVs already exceeds the towing capacity of comparable ICEVs.\\
24&BEVs typically have greater acceleration and better passing abilities.\\
25&BEVs are much quieter both in city and highway driving.\\
26&BEVs’ AWD systems provide greater low-speed control when driving off-road.\\
27&BEVs provide better weight balance which improves handling.\\
28&There is already availability of different types of BEVs, including sedans, sport cars, crossovers, trucks and minivans.\\
29&BEV owners report greater satisfaction than ICEV owners\\
30&Cheaper BEVs have much better reliability than cheaper ICEVs.\\
31&The new regulations for BEVs will require from manufacturers a 10-year / 150,000 miles warranty for batteries.\\
32&Highly reputable car companies are adding BEVs to their model lineups.\\
33&BEVs have lower center of gravity which increases stability in turns.\\
34&Greater adoption of BEVs will increase energy independence and can help national security.\\
35&New government initiatives regarding BEVs will require battery production in the USA, which will create jobs and boost the economy.\\
\hline
\end{tabular}
\end{table*}

\bibliography{aaai24}